\documentclass[10pt,twocolumn,letterpaper]{article}

\usepackage{iccv}
\usepackage{times}
\usepackage{epsfig}
\usepackage{graphicx}
\usepackage{amsmath}
\usepackage{amssymb}
\usepackage{caption}

\usepackage{booktabs}
\usepackage{multirow}
\usepackage{tabularx}
\usepackage{makecell}
\usepackage{ltablex}
\usepackage{longtable}
\usepackage{tabu}
\usepackage{tablefootnote}
\usepackage[accsupp]{axessibility}  

\makeatletter
\let\@oldmaketitle\@maketitle
\renewcommand{\@maketitle}{\@oldmaketitle
     \centering
     \vspace{-1em}
     \includegraphics[width=0.925\linewidth]{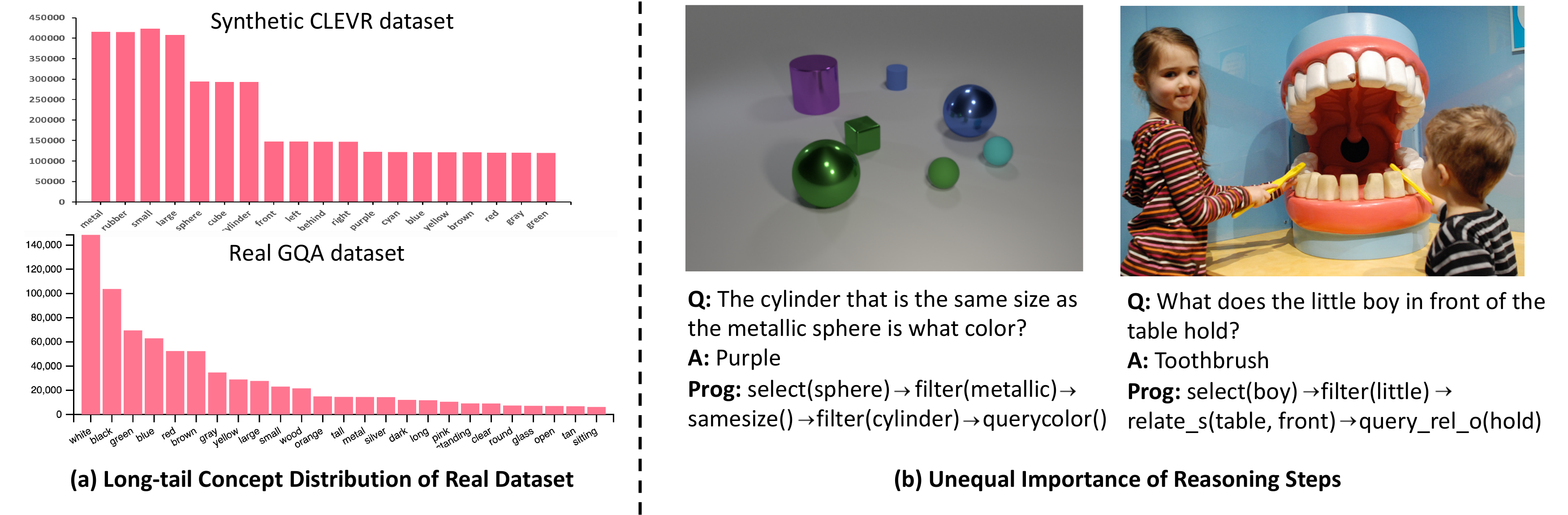}
     \vspace{-1em}
     \captionof{figure}{Statistics and examples from the synthetic CLEVR dataset and the real GQA dataset. Compared to the synthetic dataset, VQA on real data needs to deal with long-tail concept distribution and uneven importance of reasoning steps.}
     \label{fig:intro}
    \bigskip}
\makeatother

\makeatletter\renewcommand\paragraph{\@startsection{paragraph}{4}{\z@}
  {.5em \@plus1ex \@minus.2ex}{-.5em}{\normalfont\normalsize\bfseries}}\makeatother
\usepackage{hyperref}
\hypersetup{pagebackref=true,breaklinks=true,letterpaper=true,colorlinks,bookmarks=false}
\PassOptionsToPackage{linktocpage}{hyperref}

\newcommand*{\affaddr}[1]{#1} 
\newcommand*{\affmark}[1][*]{\textsuperscript{#1}}

\iccvfinalcopy 

\def\iccvPaperID{6608} 

\ificcvfinal\fi

\begin{document}

\title{Calibrating Concepts and Operations: \\ Towards Symbolic Reasoning  on  Real Images}

\author{%
Zhuowan Li\affmark[1] \quad Elias Stengel-Eskin\affmark[1] \quad Yixiao Zhang\affmark[1] \quad Cihang Xie\affmark[2] \\ Quan Tran\affmark[3] \quad Benjamin Van Durme\affmark[1]  \quad Alan Yuille\affmark[1] \vspace{.3em}\\
\affaddr{\affmark[1]Johns Hopkins University    } \qquad \quad
\affaddr{\affmark[2] University of California, Santa Cruz    } \qquad \quad
\affaddr{\affmark[3]Adobe Research}
\vspace{-.25em}
}


\makeatletter\renewcommand\paragraph{\@startsection{paragraph}{4}{\z@}
  {.5em \@plus1ex \@minus.2ex}{-.5em}{\normalfont\normalsize\bfseries}}\makeatother
\newcommand{\cihang}[1]{{\color{magenta}[cihang: #1]}}
\newcommand{\yixiao}[1]{{\color{magenta}[yixiao: #1]}}

\maketitle
\ificcvfinal\thispagestyle{empty}\fi

\begin{abstract}
   
  While neural symbolic methods demonstrate impressive performance in visual question answering on synthetic images, their performance suffers on real images. We identify that the long-tail distribution of visual concepts and unequal importance of reasoning steps in real data are the two key obstacles that limit the models' real-world potentials. To address these challenges, we propose a new paradigm, \textbf{Calibrating Concepts and Operations (CCO)}, which enables neural symbolic models to capture underlying data characteristics and to reason with hierarchical importance. Specifically, we introduce an executor with learnable concept embedding magnitudes for handling distribution imbalance, and an operation calibrator for highlighting important operations and suppressing redundant ones. 
  
  Our experiments show CCO substantially boosts the performance of neural symbolic methods on real images. By evaluating models on the real world dataset GQA, CCO helps the neural symbolic method NSCL outperforms its vanilla counterpart by 9.1\%  (from 47.0\% to 56.1\%); this result also largely reduces the performance gap between symbolic and non-symbolic methods. Additionally, we create a perturbed test set for better understanding and analyzing model performance on real images. Code is available at \url{https://github.com/Lizw14/CaliCO.git}. 
  
  
  
   
   
\end{abstract}

\vspace{-1.6em}
\section{Introduction}






Visual question answering (VQA) aims to develop a model that can answer open-ended questions from images.
Currently, end-to-end methods, which directly make predictions over dense visual and textual features \cite{yang2016stacked, kim2018bilinear}, represent the most effective class of models for VQA. Nonetheless, such methods have been criticized for exploiting shortcuts (\eg, statistical dataset bias \cite{agrawal2016analyzing, goyal2017making}, question prior \cite{agrawal2018don} or isolated text and image elements \cite{manjunatha2019explicit}) to answer questions; these shortcuts often make them unable to generalize well on out-of-domain data.

In contrast, neural symbolic methods \cite{andreas2016neural, johnson2017inferring,yi2018neural, mao2019neuro} are equipped with strong reasoning ability, enabling them to answer multi-hop and complex questions in a compositional and transparent manner---they first parse each question into a program with a series of reasoning steps, and then compose neural modules on the fly to execute the program on the image.
While symbolic methods achieve nearly perfect performance on synthetic dataset, they perform poorly on real-world datasets. For instance, neural symbolic concept learner (NSCL) \cite{mao2019neuro} achieves 98.9\% accuracy on the synthetic CLEVR dataset \cite{johnson2017clevr}, but only 47.0\% accuracy on the real-world GQA dataset \cite{hudson2019gqa}. Note the original NSCL cannot be directly applied to GQA; this 47.0\% accuracy is obtained from our own re-implementation, where minimal but necessary modifications are made (\eg, adding in $same$ and $common$ modules), for making models runnable on GQA.

As summarized in Figure \ref{fig:intro}, we note there are two major differences between 
synthetic datasets and real-world datasets. First,  while visual concepts are well-balanced in the synthetic datasets, they follow a long-tail distribution in real-world datasets. 
For example, as shown in Figure \ref{fig:intro}(a), in GQA, common concepts like ``man'', ``window'', ``black'', ``white'' are far more frequent than uncommon ones like ``pink'' and ``eraser'', in both questions and answers. Second, unlike in synthetic data, the reasoning steps on real data 
have varying importance, mainly because of redundancy/over-specification in question description. For example, as shown in Figure \ref{fig:intro}(b), in the question "What is the little boy doing?", the noun (\ie, boy) itself is enough to select the person being asked about while the adjective (\ie, little) only serves as a nuisance factor.


We identify that this mismatch of dataset characteristics is the main obstacle for adapting neural symbolic methods from synthetic datasets to real-world datasets. More concretely, we find that the original architecture designs of neural symbolic methods (which were designed/verified mainly on synthetic datasets) are no longer suitable for the real-world setting.
For examples, as shown in Section \ref{sec:motivation}, even simple operations like removing the normalization on concept embeddings or manually assigning larger weights to less discriminative modules are effective to improve the performance of neural symbolic methods on real images.


To better cope with real images, 
we propose \emph{Calibrating Concepts and Operations (CCO)}, which enables neural symbolic methods to explicitly learn weights for concept embedding and reason with contextual module importance. Specifically, CCO learns different concept embedding magnitudes for each execution module, and learns an operation weight predictor to contextually predict weights for each operation in the reasoning program. In this way, the model will be able to handle unbalanced concept distributions and to reason with varying operation importance. 

Our empirical results show that CCO substantially boosts the applicability of neural symbolic methods on real images. For example, on the real-world GQA dataset, CCO outperforms the baseline NSCL by a large margin of 9.1\% (from 47.0\% to 56.1\%).
Moreover, the proposed CCO method largely reduces the performance gap between the symbolic method and the state-of-the-art non-symbolic methods \cite{tan2019lxmert, hudson2019learning} on real-world GQA dataset. 



Additionally, based on the proposed operation weight calibrator, we create a perturbed test set by progressively removing the operations with low weights from testing questions. Our purpose is to verify whether the learned operation weights are able to highlight important operations and suppress redundant ones, and simultaneously to access the robustness of different models regarding this operation information erasing.
Our analysis reveals 
1) GQA questions contain superfluous information by way of over-specification and 2) the ability to effectively handle this extraneous information is crucial for models to improve performance. 
We hope this perturbed test set will allow researchers to better understand the compositionality of VQA questions and to further improve symbolic reasoning over real images. 


\section{Related Work}

\paragraph{Visual Question Answering} (VQA) \cite{antol2015vqa} requires an understanding of both visual and textual information. Pure deep learning methods that based on convolution, LSTM and attention have achieved good performance. For example, Fukui \etal \cite{fukui2016multimodal} used multimodal compact bilinear pooling to combine visual and language features into a joint representation. Yang \etal \cite{yang2016stacked} used stacked attention to refine the attended image region relevant to the question. Kim \etal \cite{kim2018bilinear} proposed bilinear attention network to learn attention between the two modalities with residual connections between multiple attention maps. Yang \etal \cite{yangtrrnet} proposed a tiered relational reasoning method that dynamically attends to visual objects based on textual instruction.
\paragraph{Visual reasoning.} Prior work has suggested that above mentioned VQA models may rely on dataset shortcuts and priors to predict answer \cite{agrawal2016analyzing, goyal2017making, agrawal2018don, ramakrishnan2018overcoming, cadene2019rubi, chen2020counterfactual}. 
Therefore, recent efforts have focused more on visual reasoning with complex compositional questions that requires multi-step reasoning and true visual scene understanding. Johnson \etal \cite{johnson2017clevr} propose CLEVR that requires reasoning over synthetic scenes with compositional questions automatically generated using question templates. Hudson \etal \cite{hudson2019gqa} further constructed GQA, a dataset with real images and procedurally generated multi-step questions, for visual reasoning.

\paragraph{Attention} is widely used in vision and language tasks, including image captioning \cite{xu2015show, liu2017attention, lu2017knowing, li2020context}, visual question answering \cite{yang2016stacked, kim2018bilinear, yangtrrnet}, referring expressions \cite{yu2018mattnet, yang2019dynamic, wang2019neighbourhood}. It is shown effective in learning distinct importance of images in an image group, of sub-regions over an image or of words in a sentence. Our work calibrates different concepts and operations, thus enabling the model to reason with weighted concepts and contextual operation importance. 

\paragraph{Neural symbolic methods.} \cite{zhang2019learning, zhang2021abstract, zhang2021acre} show impressive reasoning ability on abstract reasoning tasks like \cite{zhang2019raven, zhang2020number}. For VQA, Andreas \etal \cite{andreas2016neural} propose neural modular networks, which decompose a question into a functional program (reasoning steps) that can be executed by neural modules over the image. 
This method is further improved by
executing the functional program explicitly \cite{johnson2017inferring, mascharka2018transparency, hudson2018compositional, hudson2019learning, chen2021meta} or implicitly \cite{hu2017learning, hu2018explainable}, manipulating visual and textual features using convolution or dual attention. Specifically, \cite{yi2018neural, mao2019neuro, li2020competence} propose a pure symbolic executor given pre-parsed or learned explicit programs, and achieve state-of-the-art performance on CLEVR. Relatedly, Amizadeh \etal  \cite{amizadeh2020neuro} propose a symbolic reasoner based on first order logic to diagnose reasoning behavior of different models. While symbolic methods provide interpretable programs,
their reasoning capacity on real data is still limited \cite{hudson2019gqa}.
Our work aims to reduce the performance gap between symbolic and non-symbolic models on real data.


\section{Motivation} \label{sec:motivation}
In this section, we provide simple examples to demonstrate how the dataset differences (between the synthetic CLEVR and the real GQA) affect the performance of neural symbolic methods. Interestingly, we find that the traditional design principles in neural symbolic methods, which are usually obtained from synthetic datasets, may not be optimal for the real-world datasets.






\subsection{Normalized Concept Embedding?} \label{sec:motivation_oc}


For neural symbolic methods, at each step of execution, a similarity score between each object embedding and the learned concept semantic embedding is computed to select the target object that is being asked about (\ie, selecting the object that is closest to the query concept) and to predict answers (\ie, selecting the concept that is closest to the target object). By default, normalization is applied to both object embedding and concept embedding.

Interestingly, on the real-world GQA, we find this default strategy is not optimal; simply removing the normalization on concept embedding yields substantially better performance (+3.4\%). This phenomenon 
indicates that in addition to the angle alignment between object embedding and concept embedding, the magnitude of concept embedding is also informative for symbolic reasoning on real images.

We conjecture this is because the magnitude can represent the concept distribution, which is drastically different between synthetic datasets and real datasets. For example, while CLEVR contains only a relatively small and perfectly-balanced set of concepts (\ie, 19 concepts including shapes, materials), real datasets deal with thousands of concepts which are far more complex and follows a long-tail distribution. We validate this hypothesis in Section \ref{sec:analysis}---with a learnable magnitude for each concept embedding, we find its value is strongly correlated with concept frequency, \ie, more frequent concepts tend to have larger magnitudes.

\begin{figure}[t!]
\begin{center}
  \includegraphics[width=\linewidth]{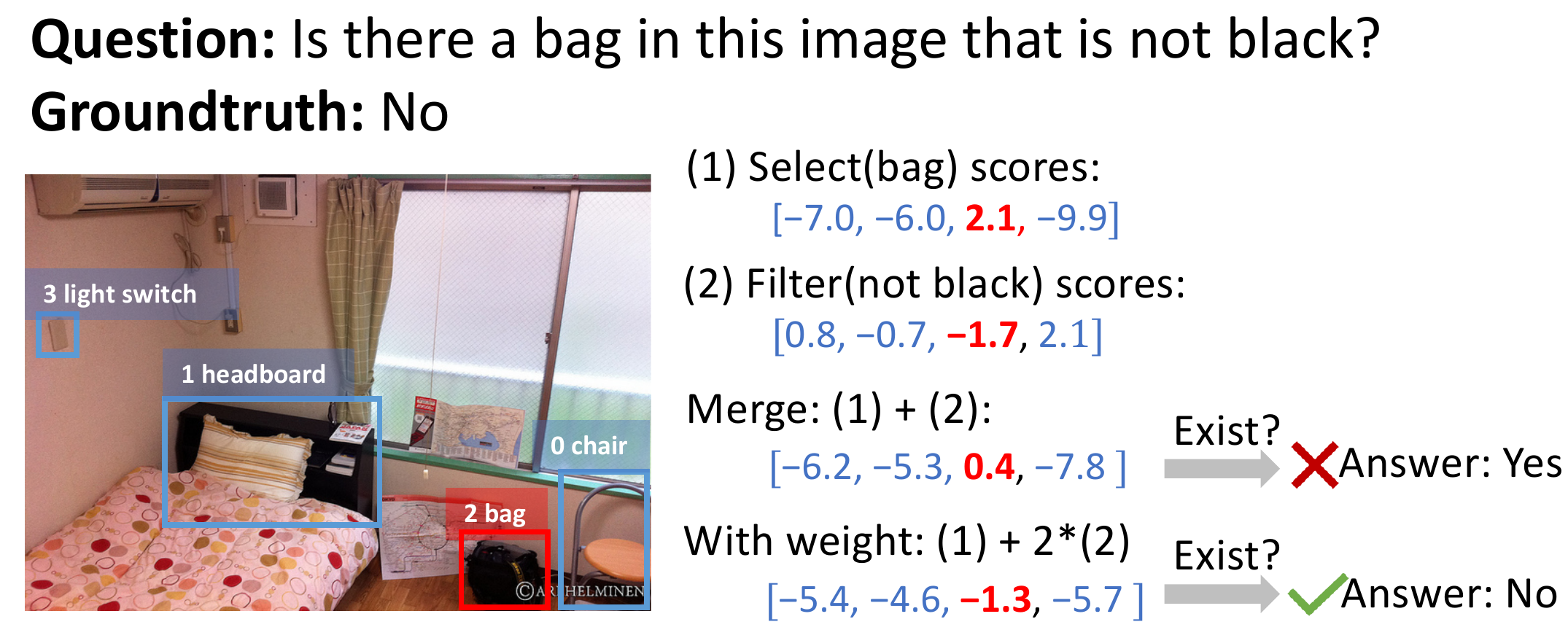}
\end{center}
\vspace{-2em}
   \caption{A failure case that can be corrected by re-weighting the operations. The $select(bag)$ operation overrides $filter(not\ black)$, thus lead to incorrect answer. This can be corrected by scaling up the result of $filter$ operation.}
\vspace{-1em}
\label{fig:motivation}
\end{figure}
\subsection{Module Re-weighting} \label{sec:motivation_ow}
In addition to this long-tailed distribution, the reasoning steps on real data are of varying importance during execution. For example, in most cases, the $select(noun)$ module are more discriminative than the $filter(attribute)$ or the $relate(relationship)$ operations, due to implicit entailment in natural language and over-specification of the question (\eg, ``little boy'', ``trees below the sky''). 
Therefore directly adapting symbolic methods to GQA will bias the model towards putting more focus on learning discriminative operations while neglecting the rest,  
resulting in errors on questions where all operations are important. 
For example, in Figure \ref{fig:motivation}, the question asks for a bag that is not black; but $select(bag)$ operation produces large values, overriding the $filter(not\ black)$ step, leading to a ``yes" answer, even though the bag is \emph{not} in the required color. 

Surprisingly, in this example, if we simply magnify the output of $filter(not\ black)$ operation by a factor of 2, the $filter$ operation then can successfully rule out the black bag, thus correctly answering the question. 
This result suggests that, while many questions contain redundant operations that the model tends to overlook, correctly re-weighting the operations is crucial for symbolic reasoning on real images. 



\section{Calibrating Concepts and Operations}

Given the observations in Section \ref{sec:motivation}, we next explore designing more sophisticated algorithms for automatically and effectively dealing with 
the complex characteristics of real data (\eg, long-tailed distribution and unequal reasoning steps), for the purpose of increasing neural symbolic methods' real-world applicability.




\subsection{Formulation}

In symbolic reasoning,  a parser first parses a question $Q = <\hat{w}_{1}, ..., \hat{w}_{l}>$ into a tree-structured functional program $P$. The program $P$ consists a set of modules $<p_{1}, ..., p_{m}>$ with dependency relationships between each other. As the functional program is either a chain or a binary tree, it can be linearized into sequence by pre-order traversal. Each operation $p$ has its type $p^{t}$ (e.g., $select$, $filter$), attribute $p^{a}$ (e.g., $color$, $material$) and concept $p^{c}$ (e.g., $red$, $plastic$). We denote the total number of module types, attributes and concepts as $n_{t}, n_{a}, n_{c}$, respectively. Then execution modules are composed on the fly, based on this generated program $P$. The module outputs are merged based on their dependency relationship and fed into the final module to get the answer $\mathbf{a}$. 

For scene representation, we first obtain a set of feature vectors $\mathbf{v}_{i} \in \mathbb{R}^{d}$ from the image $I$, with $n$ objects detected in the image. Specifically, the feature vector $\mathbf{v}$ can be either visual features obtained from Faster RCNN \cite{ren2015faster}, or the symbolic representation for each object (which can be obtained by concatenating distributions over $N_{c}$ object categories and $N_{a}$ attributes). 


\begin{figure*}[t!]
\begin{center}
\includegraphics[width=1.0\linewidth]{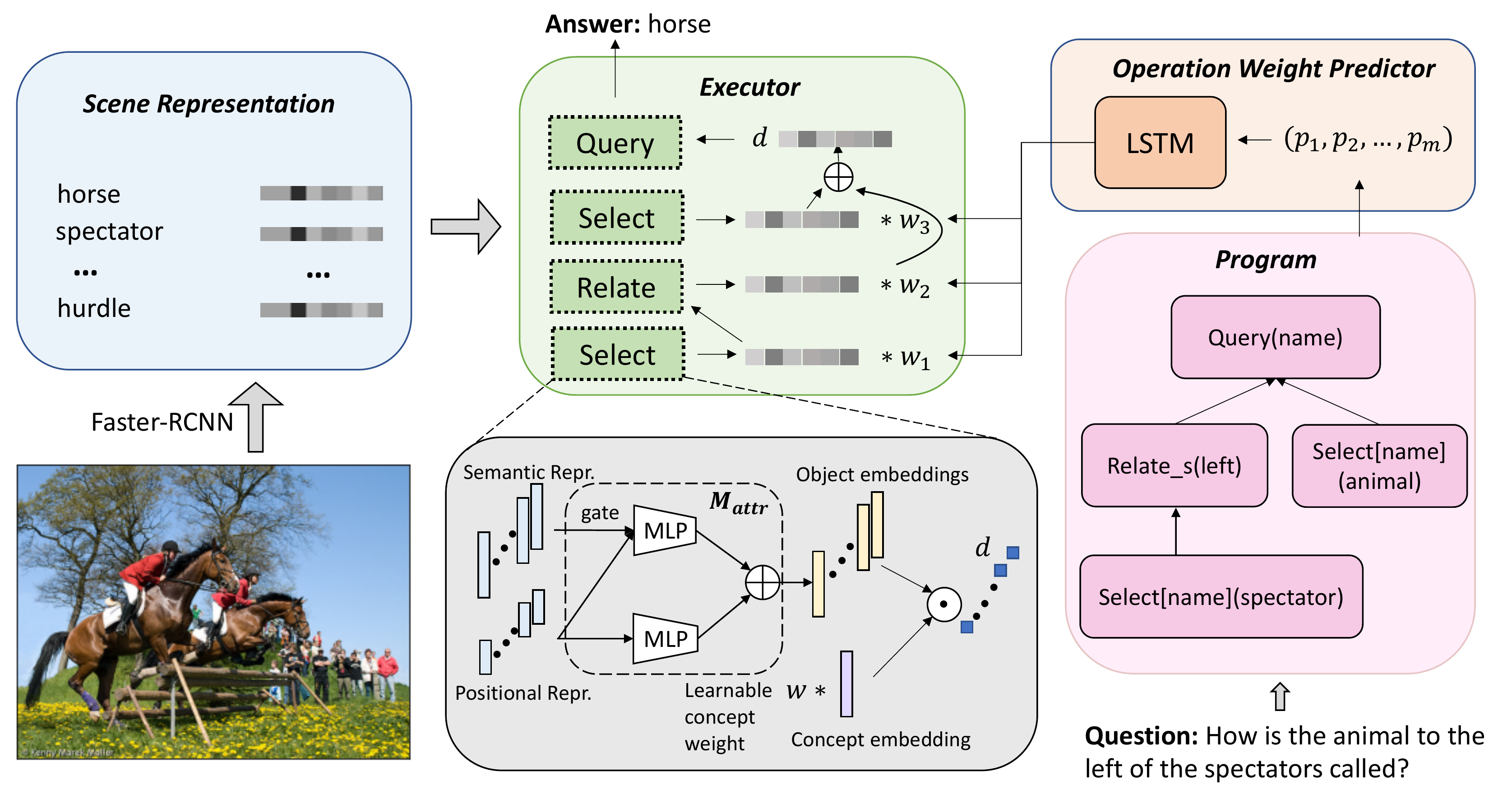}
\end{center}
\vspace{-1.5em}
   \caption{Overview of our method. We first parse the image into a symbolic scene representation in the form of objects and attributes, then parse the question into a program. In each reasoning step, a reasoning module takes in the scene representation and the instruction from the program, and outputs a distribution over objects. The Operation Weight Predictor predicts a weight for each reasoning module, which will be used to merge module outputs based on the program dependency. The final distribution is fed into the output module to predict answers.}
\vspace{-0.5em}
\label{fig:method}
\end{figure*}

\subsection{Basic Executor Architecture} \label{sec:basic_method}
Given the program $P$, the executor then executes it over input scene representations $\mathbf{v}$ to get a reasoning answer $\mathbf{a}$. The basic executor principle follows the design in \cite{mao2019neuro}.

As shown in Figure \ref{fig:method}, each module (except for the output module) produces a distribution $\mathbf{d}$ over $N$ objects in the image ($\mathbf{d} \in \mathbb{R}^{N}$), which are then merged based on their dependencies. In contrast to the default setup in the synthetic dataset, we use the mean operation (rather than minimum as in NSCL) to merge the module results due to its more stable training behavior. Finally, the output module takes in the object distribution produced by intermediate modules and queries/verifies the specified attribute of the selected object. 

For module design, a semantic embedding $\mathbf{c}$ is learned for each concept (\eg, man, red, round, etc). Without loss of generality, we illustrate the computation for $select$ and $query$ modules. The architecture of the other module types can be found in supplementary materials. 

We take the module $select[name](spectator)$ as an example. First, a small network $\mathcal{M}_{name}$ maps each object representation $\mathbf{v}_{i}$ into the concept embedding space, and then the similarity $\mathbf{s}_{i}$ between the embedded object representation $\mathbf{e}_{i}$ and the embedding of concept ``spectator'' ($\mathbf{c}_{spectator}$) is computed. This similarity $\mathbf{s}_{i}$ can be interpreted as the likelihood of each object being ``spectator". The computation of $select$ module can be summarized as the following:
\begin{align}  
\mathbf{e}_{i} &= \mathcal{M}_{attr} (\mathbf{v}_{i}) \\
s_{i} &= \operatorname{sim} (\mathbf{e}_{i}, \mathbf{c}_{cept} ) \label{eq:sim1}\\
\mathbf{d}_{select} &= [s_{1}, s_{2}, ..., s_{N}]
\end{align}
where cosine similarity, \ie, dot product of normalized $\mathbf{e}$ and $\mathbf{c}$, is used for similarity computation. 

The detailed network architecture of the representation mapping network $\mathcal{M}_{attr}$ is shown in Figure \ref{fig:method}. It gates the input object representation and passes it through a MLP to get the corresponding semantic embedding. The semantic embedding is then added with spatial embedding to get the final object embedding. The mapping networks $\mathcal{M}$ corresponding to different attributes share the same network architecture but with different weights. 

We also briefly summarize the computation of $query$ module below, as another example:

\begin{align}  
\mathbf{e}_{i} &= \mathcal{M}_{attr} (\mathbf{v}_{i}) \\
\mathbf{e} &= \mathbf{d} \cdot [\mathbf{e}_{1}, \mathbf{e}_{2}, ..., \mathbf{e}_{N}] \\
\mathbf{a}_{j} &= \operatorname{sim} (\mathbf{e}, \mathbf{c}_{j}) \label{eq:sim2}
\end{align}
where the operation $\cdot$ refers to element-wise multiplication between two vectors, and $\mathbf{c}_{j}$ refers to the concept embedding of possible answers.

\subsection{Calibrating Concepts and Operations}
We hereby formally propose \emph{Calibrating Concepts and Operations (CCO)}, which includes a concept calibration module and an operation calibration module, to help neural symbolic methods improving their applicability on real images. The overall design is illustrated in Figure \ref{fig:method}.

\paragraph{Calibrating concepts.}
As diagnosed in Section \ref{sec:motivation}, the magnitude of concept embedding $\mathbf{c}$ is informative for measuring the similarity between the object embedding and concept embedding. This motivates us to design an extra architectural element for explicitly capturing such information in magnitudes. 
Moreover, this designed architectural element is expected to be adaptive for different concepts, as each distinct type of operation is dealing with varying concept frequency distributions. For example, general concepts like ``person'' are common in $select$ module, but not in $query$ as the answers usually expect more specific concepts. 

In light of these intuitions, we offer a simple solution---explicitly learning different embedding magnitudes for each module type. We expect the learned norm sizes can encode the concept distribution, thus more frequent concepts have larger norms sizes, leading to larger similarity values. Concretely, we calibrate concept embeddings by:
\begin{align}
    \mathbf{c}_{concept} = w^{type}_{concept} \mathbf{c}_{concept} \label{eq:norm}
\end{align}
where $w$ is different for each module type and each concept. This is applied whenever concept embeddings are used for similarity computation (\eg in Equation \ref{eq:sim1}). To this end, distinct types of modules share the same concept embedding direction, but varying magnitudes, corresponding to different concept distributions.

\paragraph{Calibrating operations.}
As shown in Section \ref{sec:motivation}, on real images, it is important to enable the model to reason with different operation importance. To this end, we propose to customize the weight of each operation in the program. Specifically, a bi-directional LSTM weight predictor is used here to predict operation weights based on the whole program. For each operation $p_{i}$ in the program, its weight $w_i$ is computed as following:
\begin{align}
    \mathbf{e}_{i} &= [\mathbf{e}^{t}_{i}; \mathbf{e}^{a}_{i}; \mathbf{e}^{c}_{i}] \\
    \mathbf{h}_{1}, ..., \mathbf{h}_{m} &= \operatorname{LSTM}(\mathbf{e}_{1}, ..., \mathbf{e}_{m}) \\
    w_{i} &= \operatorname{sigmoid}(W \mathbf{h}_{i}) 
\end{align}
where $m$ is the program length. The inputs $\mathbf{e}$ to LSTM is the concatenation of the operation type embedding $\mathbf{e}^{t}$, the attribute embedding $\mathbf{e}^{a}$ and the concept embedding $\mathbf{e}^{c}$. The predicted operation weights are then used to merge outputs of the operations with a weighted-sum operation:
\begin{align}
    \mathbf{d}_{i} = \sum_{j \in \mathcal{D}(p_{i})} w_{j} \mathbf{d}_{j}
\end{align}
where $\mathcal{D}(p_{i})$ is the set of dependency operations of operation $p_{i}$. In this way, operations with higher weights play a more important role in the merging step.

\paragraph{Summary.} With the proposed CCO, neural symbolic executor now is able to capture the underlying data characteristics and reason with learnable operation importance. As we will next show, CCO substantially boosts model performance on GQA, meanwhile largely reduces the performance gap between symbolic and non-symbolic methods.

\section{Experiments} \label{sec:experiments}
\subsection{Dataset and Experiment Setup}

\paragraph{Dataset.} Our experiments are on GQA \cite{hudson2019gqa}, which is a dataset focusing on reasoning and compositional question answering over real images. Building on top of Visual Genome dataset \cite{krishna2017visual}, it contains more than 110K images and 22M questions. Each image is annotated with a scene graph cleaned from Visual Genome that contains the information of objects, attributes and relationships. Each question comes with a corresponding functional program that specifies reasoning steps. 
By default, we use its balanced version with 943k, 132k, 13k and 95k questions in train, val, testdev and test split for training and evaluating. 

\paragraph{Scene representation.} We train a Faster RCNN with an additional attribute head using cross entropy loss following \cite{anderson2018bottom}. We train with 1313 object classes (lemmatized and with plurals removed) and 622 attributes. The model gets 24.9 mAP for object detection and 17.1  groundtruth attribute average rank.\footnote{Attribute prediction is evaluated by the average rank of groundtruth attribute in all the 622 attributes. We only consider the correctly detected objects (IOU$>$0.5) for attribute evaluation.} The 1935-d concatenation of class and attribute scores are used as symbolic scene representation.

\paragraph{Implementation details.} The inner dimension of our model is 300. The concept embedding is initialized using GloVe embedding \cite{pennington2014glove}. We train our reasoning model using the Adam optimizer with an initial learning rate of 0.0005 and a batch size of 256. Linear learning rate is used with 2000 warm-up steps. We train the model for a total of 30 epochs, with early stopping (based on accuracy on the balanced testdev split) to prevent overfitting. To avoid confounding caused by parsing errors, we use \emph{gold programs} to analyze the execution performance by default. 

\subsection{Execution Results} \label{sec:execution_results}

\begin{table}[h]
\begin{center}
\begin{tabular}{l|ccc} \toprule
         &{\bf Concept}    & {\bf Operation}  & {\bf Acc.}        \\ \midrule
1 (Baseline) & Normalized     & Average     & 47.01      \\
2            & Normalized     & Calibrated    & 51.30      \\ 
3            & Unnormalized   & Calibrated    & 54.65      \\ 
{\bf 4 (Ours)}     & {\bf Calibrated} & {\bf Calibrated}    & {\bf 56.13}      \\
\bottomrule
\end{tabular}
\end{center}
\vspace{-2em}
\caption{Accuracy comparison on the balanced GQA testdev split. Compared to the baseline, both concept calibration and operation calibration substantially improve model performance. The best performance is achieved by calibrating both concept and operation.
}
\vspace{-0.5em}
\label{tab:analysis}
\end{table}

We choose NSCL \cite{mao2019neuro} as our baseline model. By  default, concept embeddings are normalized before similarity computation (cosine similarity) and operation results are merged by taking the average.  After appplying minimal but necessary changes to NSCL for making it runnable on GQA,  it achieves 47.01\% accuracy. We then integrate the proposed concept and operation calibration strategies on top of this baseline, while keeping other settings unchanged. As shown in the fourth row of Table \ref{tab:analysis}, CCO helps the baseline gain a substantial improvement, \ie, the accuracy is increased from 47.01\% to 56.13\%.
This 9.12\% improvement margin in accuracy demonstrates the effectiveness of our proposed method. 

To further analyze the improvement brought by each individual component, we progressively add in our proposed concept calibration and operation calibration into the NSCL baseline. As shown in the second row of Table \ref{tab:analysis} where the operation calibration is added, it outperforms the baseline by 4.29\%, demonstrating 
the effectiveness of operation calibration. We then remove the normalization of concept embeddings and keep the embedding magnitudes when computing similarity. As shown in the third row of Table \ref{tab:analysis}, such strategy  successfully leads to an additional 3.35\% improvement. 
This result suggests that the embedding magnitudes are informative, which is consistent with our analysis in Section \ref{sec:motivation_oc}. 
In summary, these results support that both concept weighting and operation weighting are useful for improving the NSCL baseline.

\subsection{Ablations}
\paragraph{Scene representations.}
Regarding scene representations, besides using symbolic representations, we also test model performance with other alternatives. To validate the correctness of our model design, we feed the operation modules with gold scene representation. Our CCO achieves 89.61\% accuracy, which is similar to human performance (89.30\%). This high upper bound indicates that model performance can be further improved by better visual perception.

We also examine the model performance by using visual features (Faster-RCNN feature after mean-pooling) as scene representation. Our CCO achieves 53.00\% accuracy, where the 3.13\% performance gap (\ie, 53.00\% \vs 56.13\%) shows the advantage of the abstract symbolic scene representation over the dense visual features.


\paragraph{Program parsing.}
In all previous experiments, we apply gold program for facilitating performance analysis. While in this part, we now examine the model performance in the wild, \ie, gold program is no longer available.  In order to parse the question into functional program, we apply MISO, a popular sequence-to-graph parser used for parsing in a number of graph-based formalisms \cite{zhang2019amr, zhang2019broad, stengel2020universal}. Different from simple sequence-to-sequence parser as in \cite{johnson2017inferring} that can only handle program with one argument, or the two-stage parser as in \cite{chen2021meta} that handles multiple arguments by hard constraints, MISO can automatically handle multiple arguments by treating the program as a graph. The inputs to the MISO parser are word embedding sequences and output is a pre-order traversal of a program trees.

We present the parsing results in Table \ref{tab:parsing}.  We use exact match score, which is calculated by the percentage of predicted programs that exactly match the gold program, for measuring the quality of the predicted program. Our parser outperforms the parser in MMN \cite{chen2021meta} by a large margin of 6.05\% in terms of exact match score. Nonetheless, interestingly, we find final model accuracy is less impacted by the quality of program---by executing either ours or MMN's predicted program, the difference in the final model accuracy is only 0.1\%. This seemly ``frustrating'' result may suggest the performance of other components in current neural symbolic methods are severely lagged behind therefore are not able to cope with the advances brought by our strong parser.

\begin{table}[h]
    \centering
    \begin{tabular}{l|cc}
    \toprule
    Model & Exact match & Acc.\\
    \hline
        MMN \cite{chen2021meta} & 85.13 & 54.01 \\
        Ours & \textbf{91.18} & \textbf{54.11} \\
    \bottomrule
    \end{tabular}
    \vspace{-1em}
    \caption{Parsing performance on testdev\_balanced, measured by exact match score and execution accuracy.}
    \vspace{-0.7em}
    \label{tab:parsing}
\end{table}




\begin{table*}[t!]
\begin{center}
\begin{tabular}{llccccccc}
\toprule
    & {\bf Method}   & {\bf Acc}   & {\bf Binary} & {\bf Open}  & {\bf Consistency} & {\bf Plausibility} & {\bf Validity} & {\bf Distribution} \\ \midrule
\multirow{3}{*}{Non-Symbolic} & LXMERT \cite{tan2019lxmert} & 60.33 & 77.16  & 45.47 & 89.59       & 84.53        & 96.35    & 5.49         \\
 & NSM \cite{hudson2019learning}   & 63.17 & 78.94  & 49.25 & 93.25       & 84.28        & 96.41    & 3.71         \\
 & MMN \cite{chen2021meta} & 60.83 & 78.90 & 44.89 & 92.49 & 84.55 & 96.19 & 5.54 \\ \hline
\multirow{2}{*}{Symbolic}  & $\nabla$-FOL  \cite{amizadeh2020neuro}  & 54.76 & 71.99  & 41.22 & 84.48       & -            & -        & -            \\ 
 & \textbf{CCO (ours)}   &  56.38 & 74.83  & 40.09 & 91.71       &  83.76       &  95.43   &   6.32       \\
\bottomrule
\end{tabular}
\end{center}
\vspace{-2em}
\caption{Comparison with state-of-the-art symbolic and non-symbolic methods on the official testing split. }
\vspace{-1em}
\label{tab:performance}
\end{table*}

\paragraph{Comparing to the state-of-the-arts.}
To fairly compare different methods on GQA, we follow the training setups in \cite{amizadeh2020neuro,chen2021meta} where  we first train the model on unbalanced training split then finetuned on balanced training split. Gold programs are used for training while parser predicted programs are used for evaluation. Performance is reported using the official evaluation metrics, including overall accuracy, accuracy on binary questions, accuracy on open questions, consistency, plausibility, validity and distribution. 

We consider three non-symbolic methods (\ie, LXMERT \cite{tan2019lxmert}, NSM \cite{hudson2019learning}, MMN \cite{chen2021meta}) and one symbolic method (\ie, $\nabla$-FOL \cite{amizadeh2020neuro}) for performance comparison. In short, LXMERT is a representative multi-modal pretraining method; NSM is a graph-based model that achieves state-of-the-art performance on GQA; MMN is a modular method but is still based on dense features manipulation; $\nabla$-FOL{\footnote {$\nabla$-FOL does not report full result on the official test split, therefore results on balanced testdev split is shown for comparison.}} is a symbolic method based on first order logic and contextual calibration. 
We summarize the model performance on the held-out test split in Table \ref{tab:performance}.



Compared to the previous state-of-the-art symbolic method $\nabla$-FOL, our proposed CCO surpasses it by 1.58\% in terms of accuracy. Moreover, as shown in Table \ref{tab:performance}, we note the performance gain over $\nabla$-FOL is mainly on the binary questions (+2.84) and on predicting consistent answers for different questions (+7.2\%).


We next compare with the state-of-the-art non-symbolic methods. Though our model still has lower accuracy than these non-symbolic methods, we note their performance on consistency, plausibility and validity is on a par with each other. We conjecture this
is due to the symbolic nature of our model, \ie, the proposed CCO execute strictly according to the program, thus answers are plausible and valid, and questions with same underlying program get consistent answer. These results suggest that the proposed CCO largely reduces the performance gap between symbolic and non-symbolic methods on the real-world GQA dataset.




\section{Analysis} \label{sec:analysis}
\subsection{Learned Embedding Magnitudes}

To verify our motivation that the learned concept embedding magnitudes are informative for representing the unbalanced concept distribution in real dataset, we visualize the correlation between concept counts and their magnitude after calibration (in $query$ module), \ie, $\Vert{\mathbf{c}_{concept}}\Vert_{2}$ after calibration in Equation \ref{eq:norm}. In the plot, X-axis is the count of concepts in $query$ module (taking log), and Y-axis is the learned magnitude of concept embeddings.

\begin{figure}[t!]
\begin{center}
\includegraphics[width=0.9\linewidth]{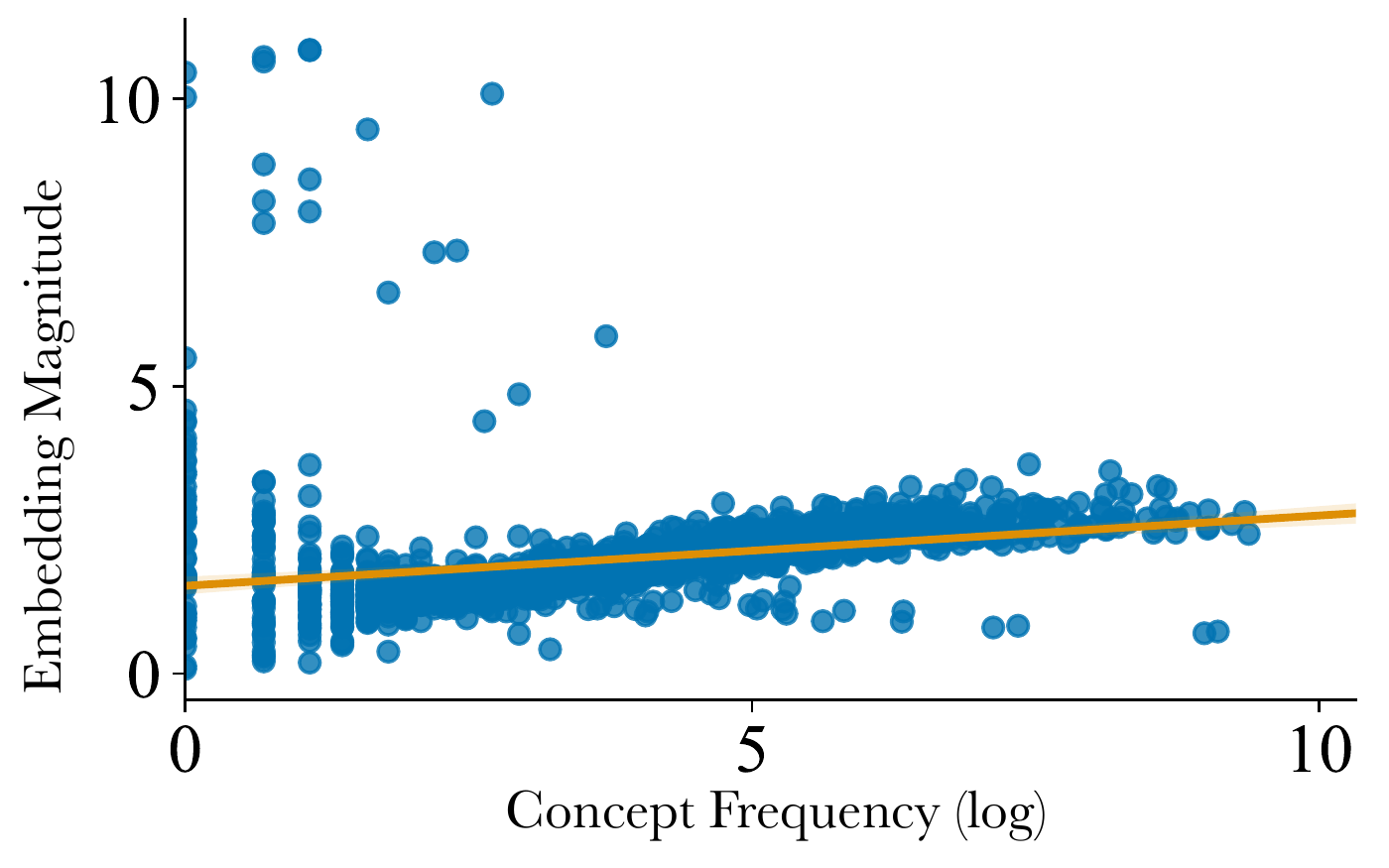}
\end{center}
\vspace{-2em}
   \caption{A positive correlation between learned embedding magnitude and concept frequency confirms our motivating intuition: more frequent concepts have larger magnitudes.}
  \vspace{-0.6em}
\label{fig:norm}
\end{figure}

As verified in Figure \ref{fig:norm}, more frequent concepts consistently learn larger magnitudes, while less frequent concepts generally have smaller magnitudes. With larger magnitudes, the frequent concepts will produce values with higher confidence when computing similarity in the output of each module. Another interesting observation is that the magnitudes for few-shot concepts are not very consistent (\ie, have larger variance), which is possibly caused by the insufficient number of training examples.

\subsection{Perturbed Test Set} \label{sec:perturb}


We create a perturbed testing data splits for the following purposes: a) we want to validate that proposed operation weighting strategy predicts larger weights for more important operations and smaller weights for unimportant ones; b) we need a test set for better studying the question over-specification in GQA dataset; and c) we aim to benchmark behavior of symbolic and non-symbolic methods in terms of how much information in the over-specified operations can be effectively utilized. 

Specifically, this perturbed test set is created using the operation weights predicted by the learned LSTM operation weight predictor. We perturb the functional programs in balanced testdev splits by progressively removing the removable operations with smaller predicted weights\footnote{We set the weight thresholds to be $-\infty$, $-2$, $-1$, $-0.5$, $0$, $+\infty$; resulting in removing $0\%$, $14\%, $31\%$, $70\%, $90\%$, $100\%$ of removable operations, respectively}. Note that removable operations here refer to the intermediate operations that can be removed without syntactically breaking the programs, \ie, $filter$, $relate$ and their dependent operations. Then, we train a simple sequence-to-sequence generator to recover questions from the perturbed programs.

\begin{figure}[t!]
\begin{center}
\includegraphics[width=0.9\linewidth]{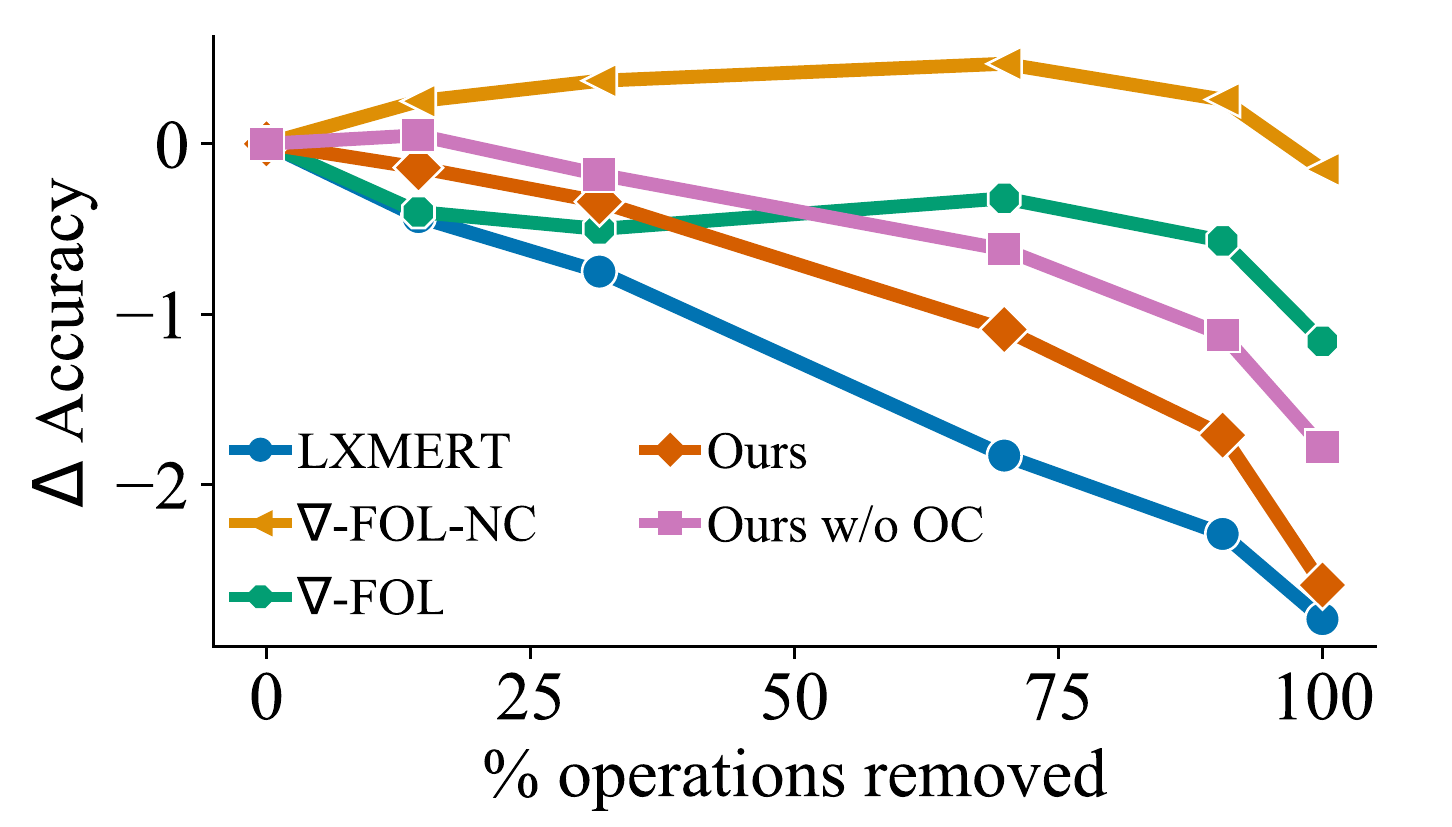}
\end{center}
\vspace{-2em}
   \caption{Accuracy drop of different models when the testing questions are progressively perturbed by removing reasoning operations with low weights.}
 \vspace{-1em}
\label{fig:perturb}
\end{figure}

The results are shown in Figure \ref{fig:perturb}. We test five methods, including non-symbolic method LXMERT \cite{tan2019lxmert}, symbolic method $\nabla$-FOL \cite{amizadeh2020neuro}, its variant $\nabla$-FOL-NC which is a pure reasoner based on first order logic, our model, and ours without operation calibrating.\footnote{Original accuracy of the five models (LXMERT, $\nabla$-FOL, $\nabla$-FOL-NC, ours, and ours w/o OC) are 58.13, 54.02, 51.86, 56.13, 55.49, respectively.} Our observations can be summarized as the following:

\paragraph{Validity of operation weights.} All curves exhibit a sharper decrease at the end when more operations with higher weights are removed. In other words, the removal of operations with larger predicted weights will result in bigger negative influence on model accuracy. This validates the predicted weights correctly represent operation importance. 

\paragraph{Question over-specification.} From the curves, we note while 59.0\% questions in the balanced testdev split contain removable operations and are perturbed, less then 3.0\% questions are incorrectly answered after removing those modules. This phenomenon suggests that for most questions in GQA dataset, the $filter$ and $relate$ operations are not necessary for figuring out the answer, \ie, removing all the intermediate attributes and relationships from questions does not change the answer for most of the questions.

\paragraph{Effectiveness of operation weighting.} Interestingly, the performance of the pure logic reasoner $\nabla$-FOL-NC and our model without operation weighting sees a slight increase when removing a small amount of operations. This phenomenon indicates that those operations are hard for models to learn thus can even derail the model predictions. This verifies our motivation for designing operation calibration as it helps the learning of $filter$ and $relate$ modules.

\paragraph{Comparison of symbolic and non-symbolic methods.} Compared to symbolic methods, non-symbolic methods have larger accuracy drops, therefore indicating they can more effectively utilize the information in adjectives and relationships. Moreover, methods with higher performance tend to have larger decrease when questions are perturbed. This suggests enhancing the model's ability to understand filtering adjectives and relationships is crucial for improving symbolic methods on real images.

\subsection{Hard and Easy Subset}
We additionally perturb the visual-hard and the visual-easy testing splits \cite{amizadeh2020neuro} and evaluate our CCO model on them. 
Specifically, the easy split contains questions that visually easy thus can be answered correctly by their differentiable first-order logic formula, while the hard split are harder in perception. 
In other words, the easy split contains questions that can be answered by a perception system alone, while the hard split contains images requiring more reasoning.
With perturbed versions, we can investigate to what degree low-weight operations are implicated in multi-step reasoning for visually hard questions.

We summarize the model performance in Table \ref{tab:easy}. 
With more operations get removed, the accuracy drop on perturbed hard split is much larger than the easy split. 
This indicates that the visually hard questions force the model to better utilize every piece information in the question, while easy questions contain more redundant operations that are not necessarily needed.

\begin{table}[t!]
\begin{center}
\begin{tabular}{c|c|cc}
\toprule
{\bf threshold}        & {\bf All}   & {\bf Easy}  & {\bf Hard}  \\ \midrule
$-\infty$(orig)    & 56.13 & 78.03     & 37.42     \\ 
-2  & -0.14 & 0.4   & -2    \\
-1  & -0.34 & 0.13  & -2.17 \\
0.5 & -1.09 & -0.51 & -3.04 \\
0    & -1.71 & -0.93 & -3.86 \\
$+\infty$        & -2.59 & -1.88 & -4.72 \\
\bottomrule
\end{tabular}
\vspace{-2em}
\end{center}
\caption{Model accuracy on perturbed easy/hard splits.}
\vspace{-1em}
\label{tab:easy}
\end{table}

\section{Conclusion} 
\vspace{-0.4em}
To improve symbolic reasoning for VQA on real images, we propose to calibrate concepts and operations (CCO), which helps models handle the unbalanced concept distribution and unequal importance of reasoning operations. Experimental results demonstrate the effectiveness of the proposed method, where CCO outperforms several baselines by a large margin and reduces the performance gap between symbolic and non-symbolic methods. Additionally, we propose a perturbed test set for better understanding and analyzing model performance on real images. We hope this dataset can help  researchers to further study the potential of symbolic reasoning on real images in the future.


\vspace{-0.4em}
\section*{Acknowledgements}

\vspace{-0.4em}
This work was supported by NSF \#1763705 and IARPA BETTER (2019-19051600005). Elias Stengel-Eskin is supported by an NSF Graduate Research Fellowship. Cihang Xie is supported by a gift grant from Open Philanthropy.

\clearpage
\newpage
{\small
\bibliographystyle{plain}
\bibliography{egbib}

\begin{thebibliography}{10}

\bibitem{agrawal2016analyzing}
Aishwarya Agrawal, Dhruv Batra, and Devi Parikh.
\newblock Analyzing the behavior of visual question answering models.
\newblock {\em arXiv preprint arXiv:1606.07356}, 2016.

\bibitem{agrawal2018don}
Aishwarya Agrawal, Dhruv Batra, Devi Parikh, and Aniruddha Kembhavi.
\newblock Don't just assume; look and answer: Overcoming priors for visual
  question answering.
\newblock In {\em Proceedings of the IEEE Conference on Computer Vision and
  Pattern Recognition}, pages 4971--4980, 2018.

\bibitem{amizadeh2020neuro}
Saeed Amizadeh, Hamid Palangi, Alex Polozov, Yichen Huang, and Kazuhito
  Koishida.
\newblock Neuro-symbolic visual reasoning: Disentangling ``{V}isual" from
  ``{R}easoning".
\newblock In Hal~Daumé III and Aarti Singh, editors, {\em Proceedings of the
  37th International Conference on Machine Learning}, volume 119 of {\em
  Proceedings of Machine Learning Research}, pages 279--290. PMLR, 13--18 Jul
  2020.

\bibitem{anderson2018bottom}
Peter Anderson, Xiaodong He, Chris Buehler, Damien Teney, Mark Johnson, Stephen
  Gould, and Lei Zhang.
\newblock Bottom-up and top-down attention for image captioning and visual
  question answering.
\newblock In {\em Proceedings of the IEEE conference on computer vision and
  pattern recognition}, pages 6077--6086, 2018.

\bibitem{andreas2016neural}
Jacob Andreas, Marcus Rohrbach, Trevor Darrell, and Dan Klein.
\newblock Neural module networks.
\newblock In {\em Proceedings of the IEEE conference on computer vision and
  pattern recognition}, pages 39--48, 2016.

\bibitem{antol2015vqa}
Stanislaw Antol, Aishwarya Agrawal, Jiasen Lu, Margaret Mitchell, Dhruv Batra,
  C~Lawrence Zitnick, and Devi Parikh.
\newblock Vqa: Visual question answering.
\newblock In {\em Proceedings of the IEEE international conference on computer
  vision}, pages 2425--2433, 2015.

\bibitem{cadene2019rubi}
Remi Cadene, Corentin Dancette, Hedi Ben-Younes, Matthieu Cord, and Devi
  Parikh.
\newblock Rubi: Reducing unimodal biases in visual question answering.
\newblock {\em arXiv preprint arXiv:1906.10169}, 2019.

\bibitem{chen2020counterfactual}
Long Chen, Xin Yan, Jun Xiao, Hanwang Zhang, Shiliang Pu, and Yueting Zhuang.
\newblock Counterfactual samples synthesizing for robust visual question
  answering.
\newblock In {\em Proceedings of the IEEE/CVF Conference on Computer Vision and
  Pattern Recognition}, pages 10800--10809, 2020.

\bibitem{chen2021meta}
Wenhu Chen, Zhe Gan, Linjie Li, Yu~Cheng, William Wang, and Jingjing Liu.
\newblock Meta module network for compositional visual reasoning.
\newblock In {\em Proceedings of the IEEE/CVF Winter Conference on Applications
  of Computer Vision}, pages 655--664, 2021.

\bibitem{fukui2016multimodal}
Akira Fukui, Dong~Huk Park, Daylen Yang, Anna Rohrbach, Trevor Darrell, and
  Marcus Rohrbach.
\newblock Multimodal compact bilinear pooling for visual question answering and
  visual grounding.
\newblock {\em arXiv preprint arXiv:1606.01847}, 2016.

\bibitem{goyal2017making}
Yash Goyal, Tejas Khot, Douglas Summers-Stay, Dhruv Batra, and Devi Parikh.
\newblock Making the v in vqa matter: Elevating the role of image understanding
  in visual question answering.
\newblock In {\em Proceedings of the IEEE Conference on Computer Vision and
  Pattern Recognition}, pages 6904--6913, 2017.

\bibitem{hu2018explainable}
Ronghang Hu, Jacob Andreas, Trevor Darrell, and Kate Saenko.
\newblock Explainable neural computation via stack neural module networks.
\newblock In {\em Proceedings of the European conference on computer vision
  (ECCV)}, pages 53--69, 2018.

\bibitem{hu2017learning}
Ronghang Hu, Jacob Andreas, Marcus Rohrbach, Trevor Darrell, and Kate Saenko.
\newblock Learning to reason: End-to-end module networks for visual question
  answering.
\newblock In {\em Proceedings of the IEEE International Conference on Computer
  Vision}, pages 804--813, 2017.

\bibitem{hudson2018compositional}
Drew~A Hudson and Christopher~D Manning.
\newblock Compositional attention networks for machine reasoning.
\newblock {\em arXiv preprint arXiv:1803.03067}, 2018.

\bibitem{hudson2019gqa}
Drew~A Hudson and Christopher~D Manning.
\newblock Gqa: A new dataset for real-world visual reasoning and compositional
  question answering.
\newblock In {\em Proceedings of the IEEE/CVF Conference on Computer Vision and
  Pattern Recognition}, pages 6700--6709, 2019.

\bibitem{hudson2019learning}
Drew~A Hudson and Christopher~D Manning.
\newblock Learning by abstraction: The neural state machine.
\newblock {\em arXiv preprint arXiv:1907.03950}, 2019.

\bibitem{johnson2017clevr}
Justin Johnson, Bharath Hariharan, Laurens Van Der~Maaten, Li~Fei-Fei,
  C~Lawrence~Zitnick, and Ross Girshick.
\newblock Clevr: A diagnostic dataset for compositional language and elementary
  visual reasoning.
\newblock In {\em Proceedings of the IEEE Conference on Computer Vision and
  Pattern Recognition}, pages 2901--2910, 2017.

\bibitem{johnson2017inferring}
Justin Johnson, Bharath Hariharan, Laurens Van Der~Maaten, Judy Hoffman,
  Li~Fei-Fei, C~Lawrence~Zitnick, and Ross Girshick.
\newblock Inferring and executing programs for visual reasoning.
\newblock In {\em Proceedings of the IEEE International Conference on Computer
  Vision}, pages 2989--2998, 2017.

\bibitem{kim2018bilinear}
Jin-Hwa Kim, Jaehyun Jun, and Byoung-Tak Zhang.
\newblock Bilinear attention networks.
\newblock {\em arXiv preprint arXiv:1805.07932}, 2018.

\bibitem{krishna2017visual}
Ranjay Krishna, Yuke Zhu, Oliver Groth, Justin Johnson, Kenji Hata, Joshua
  Kravitz, Stephanie Chen, Yannis Kalantidis, Li-Jia Li, David~A Shamma, et~al.
\newblock Visual genome: Connecting language and vision using crowdsourced
  dense image annotations.
\newblock {\em International journal of computer vision}, 123(1):32--73, 2017.

\bibitem{li2020competence}
Qing Li, Siyuan Huang, Yining Hong, and Song-Chun Zhu.
\newblock A competence-aware curriculum for visual concepts learning via
  question answering.
\newblock In {\em European Conference on Computer Vision}, pages 141--157.
  Springer, 2020.

\bibitem{li2020context}
Zhuowan Li, Quan Tran, Long Mai, Zhe Lin, and Alan~L Yuille.
\newblock Context-aware group captioning via self-attention and contrastive
  features.
\newblock In {\em Proceedings of the IEEE/CVF Conference on Computer Vision and
  Pattern Recognition}, pages 3440--3450, 2020.

\bibitem{liu2017attention}
Chenxi Liu, Junhua Mao, Fei Sha, and Alan Yuille.
\newblock Attention correctness in neural image captioning.
\newblock In {\em Proceedings of the AAAI Conference on Artificial
  Intelligence}, volume~31, 2017.

\bibitem{lu2017knowing}
Jiasen Lu, Caiming Xiong, Devi Parikh, and Richard Socher.
\newblock Knowing when to look: Adaptive attention via a visual sentinel for
  image captioning.
\newblock In {\em Proceedings of the IEEE conference on computer vision and
  pattern recognition}, pages 375--383, 2017.

\bibitem{manjunatha2019explicit}
Varun Manjunatha, Nirat Saini, and Larry~S Davis.
\newblock Explicit bias discovery in visual question answering models.
\newblock In {\em Proceedings of the IEEE/CVF Conference on Computer Vision and
  Pattern Recognition}, pages 9562--9571, 2019.

\bibitem{mao2019neuro}
Jiayuan Mao, Chuang Gan, Pushmeet Kohli, Joshua~B Tenenbaum, and Jiajun Wu.
\newblock The neuro-symbolic concept learner: Interpreting scenes, words, and
  sentences from natural supervision.
\newblock {\em arXiv preprint arXiv:1904.12584}, 2019.

\bibitem{mascharka2018transparency}
David Mascharka, Philip Tran, Ryan Soklaski, and Arjun Majumdar.
\newblock Transparency by design: Closing the gap between performance and
  interpretability in visual reasoning.
\newblock In {\em Proceedings of the IEEE conference on computer vision and
  pattern recognition}, pages 4942--4950, 2018.

\bibitem{pennington2014glove}
Jeffrey Pennington, Richard Socher, and Christopher~D Manning.
\newblock Glove: Global vectors for word representation.
\newblock In {\em Proceedings of the 2014 conference on empirical methods in
  natural language processing (EMNLP)}, pages 1532--1543, 2014.

\bibitem{ramakrishnan2018overcoming}
Sainandan Ramakrishnan, Aishwarya Agrawal, and Stefan Lee.
\newblock Overcoming language priors in visual question answering with
  adversarial regularization.
\newblock {\em arXiv preprint arXiv:1810.03649}, 2018.

\bibitem{ren2015faster}
Shaoqing Ren, Kaiming He, Ross Girshick, and Jian Sun.
\newblock Faster r-cnn: Towards real-time object detection with region proposal
  networks.
\newblock {\em arXiv preprint arXiv:1506.01497}, 2015.

\bibitem{stengel2020universal}
Elias Stengel-Eskin, Aaron~Steven White, Sheng Zhang, and Benjamin Van~Durme.
\newblock Universal decompositional semantic parsing.
\newblock In {\em Proceedings of the 58th Annual Meeting of the Association for
  Computational Linguistics}, pages 8427--8439, 2020.

\bibitem{tan2019lxmert}
Hao Tan and Mohit Bansal.
\newblock Lxmert: Learning cross-modality encoder representations from
  transformers.
\newblock {\em arXiv preprint arXiv:1908.07490}, 2019.

\bibitem{wang2019neighbourhood}
Peng Wang, Qi~Wu, Jiewei Cao, Chunhua Shen, Lianli Gao, and Anton van~den
  Hengel.
\newblock Neighbourhood watch: Referring expression comprehension via
  language-guided graph attention networks.
\newblock In {\em Proceedings of the IEEE/CVF Conference on Computer Vision and
  Pattern Recognition}, pages 1960--1968, 2019.

\bibitem{xu2015show}
Kelvin Xu, Jimmy Ba, Ryan Kiros, Kyunghyun Cho, Aaron Courville, Ruslan
  Salakhudinov, Rich Zemel, and Yoshua Bengio.
\newblock Show, attend and tell: Neural image caption generation with visual
  attention.
\newblock In {\em International conference on machine learning}, pages
  2048--2057. PMLR, 2015.

\bibitem{yang2019dynamic}
Sibei Yang, Guanbin Li, and Yizhou Yu.
\newblock Dynamic graph attention for referring expression comprehension.
\newblock In {\em Proceedings of the IEEE/CVF International Conference on
  Computer Vision}, pages 4644--4653, 2019.

\bibitem{yangtrrnet}
Xiaofeng Yang, Guosheng Lin, Fengmao Lv, and Fayao Liu.
\newblock Trrnet: Tiered relation reasoning for compositional visual question
  answering.

\bibitem{yang2016stacked}
Zichao Yang, Xiaodong He, Jianfeng Gao, Li~Deng, and Alex Smola.
\newblock Stacked attention networks for image question answering.
\newblock In {\em Proceedings of the IEEE conference on computer vision and
  pattern recognition}, pages 21--29, 2016.

\bibitem{yi2018neural}
Kexin Yi, Jiajun Wu, Chuang Gan, Antonio Torralba, Pushmeet Kohli, and Joshua~B
  Tenenbaum.
\newblock Neural-symbolic vqa: Disentangling reasoning from vision and language
  understanding.
\newblock {\em arXiv preprint arXiv:1810.02338}, 2018.

\bibitem{yu2018mattnet}
Licheng Yu, Zhe Lin, Xiaohui Shen, Jimei Yang, Xin Lu, Mohit Bansal, and
  Tamara~L Berg.
\newblock Mattnet: Modular attention network for referring expression
  comprehension.
\newblock In {\em Proceedings of the IEEE Conference on Computer Vision and
  Pattern Recognition}, pages 1307--1315, 2018.

\bibitem{zhang2019raven}
Chi Zhang, Feng Gao, Baoxiong Jia, Yixin Zhu, and Song-Chun Zhu.
\newblock Raven: A dataset for relational and analogical visual reasoning.
\newblock In {\em Proceedings of the IEEE Conference on Computer Vision and
  Pattern Recognition (CVPR)}, 2019.

\bibitem{zhang2021acre}
Chi Zhang, Baoxiong Jia, Mark Edmonds, Song-Chun Zhu, and Yixin Zhu.
\newblock Acre: Abstract causal reasoning beyond covariation.
\newblock In {\em Proceedings of the IEEE Conference on Computer Vision and
  Pattern Recognition (CVPR)}, 2021.

\bibitem{zhang2019learning}
Chi Zhang, Baoxiong Jia, Feng Gao, Yixin Zhu, Hongjing Lu, and Song-Chun Zhu.
\newblock Learning perceptual inference by contrasting.
\newblock In {\em Advances in Neural Information Processing Systems (NeurIPS)},
  2019.

\bibitem{zhang2021abstract}
Chi Zhang, Baoxiong Jia, Song-Chun Zhu, and Yixin Zhu.
\newblock Abstract spatial-temporal reasoning via probabilistic abduction and
  execution.
\newblock In {\em Proceedings of the IEEE/CVF Conference on Computer Vision and
  Pattern Recognition}, pages 9736--9746, 2021.

\bibitem{zhang2019amr}
Sheng Zhang, Xutai Ma, Kevin Duh, and Benjamin Van~Durme.
\newblock Amr parsing as sequence-to-graph transduction.
\newblock In {\em Proceedings of the 57th Annual Meeting of the Association for
  Computational Linguistics}, pages 80--94, 2019.

\bibitem{zhang2019broad}
Sheng Zhang, Xutai Ma, Kevin Duh, and Benjamin Van~Durme.
\newblock Broad-coverage semantic parsing as transduction.
\newblock In {\em Proceedings of the 2019 Conference on Empirical Methods in
  Natural Language Processing and the 9th International Joint Conference on
  Natural Language Processing (EMNLP-IJCNLP)}, pages 3777--3789, 2019.

\bibitem{zhang2020number}
Wenhe Zhang, Chi Zhang, Yixin Zhu, and Song-Chun Zhu.
\newblock Machine number sense: A dataset of visual arithmetic problems for
  abstract and relational reasoning.
\newblock In {\em AAAI Conference on Artificial Intelligence (AAAI)}, 2020.

\end{thebibliography}
}

\clearpage
\onecolumn
\appendix

\vspace{-2.4em}

\section{Execution Examples}

\begin{figure}[h]
\begin{center}
\vspace{-0.8em}
\includegraphics[width=0.95\linewidth]{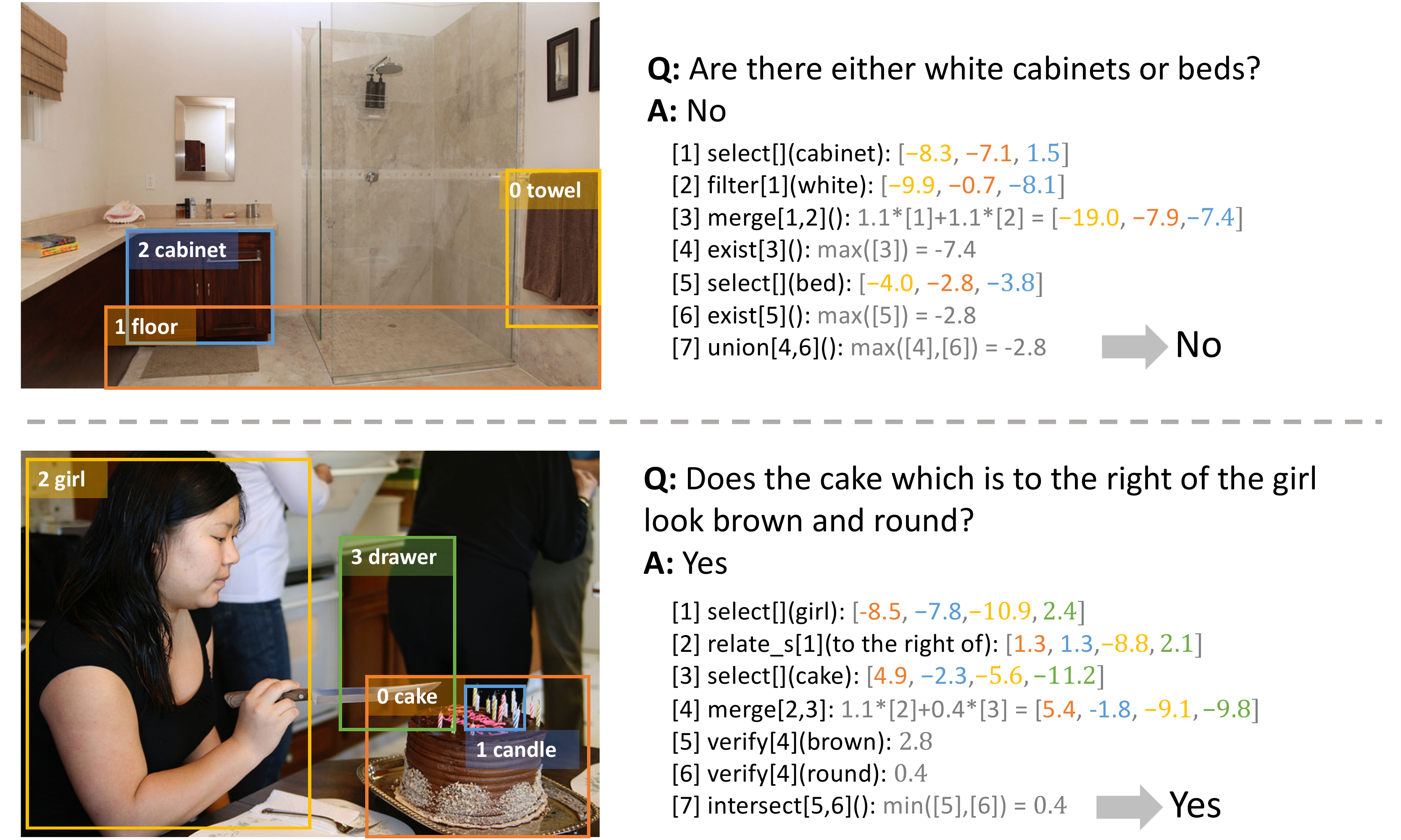}
\vspace{-1.0em}
\end{center}
   \caption{Examples of the execution process. Best view in color.}
\label{fig:supp_examples}
\vspace{-0.2em}
\end{figure}

In Figure \ref{fig:supp_examples}, two examples are shown to help better understand the reasoning process. For simplification, in each example, only a few representative object region proposal boxes are shown. The output of each execution step (shown in different colors corresponding to the image region colors) are the scores representing each region box being selected or not. By looking at the intermediate outputs, we can see how each execution step changes the selection scores. For example, in example (a), the $select(cabinet)$ step produces positive scores for the cabinet object region and negative scores for the others, but after $filter(white)$ and the $merge$ step, all scores become negative, which indicates that there is not white cabinet in this image. At last, the output modules does logistic operation on top of the final selection scores to produce answers. For example, $exist$ checks whether there is a positive score to find if there is some objects being selected.

\section{Module Details}

While Section 4.2 shows computations for $select$ and $query$ modules, here in Table \ref{tab:modules}, we show details of all modules, including their inputs, output, arguments, description and computation. In the inputs and output, bold symbols (\eg, $\mathbf{d}, \mathbf{a}$) represent vectors while plain symbols (\eg, $a$) represent scalar scores. \footnote{In training, cross entropy loss is used for open questions (where the output module produces a vector $\mathbf{a}$), while binary cross entropy loss is used for binary questions (where the output module produces a scalar score $a$.}

\begin{table}[h]
\caption{List of all modules. Inputs, output, arguments, description and computation details are shown for each module.} \label{tab:modules} 
\vspace{-0.5em}
\resizebox{1.\linewidth}{!}{
\begin{tabular}{l|l|l|l|l|c}
\toprule
{\bf Module}         & {\bf Inputs} & {\bf Output} & {\bf Arguments} & {\bf Description}  & {\bf Computation} \\ \midrule
\multicolumn{6}{c}{{\bf Intermediate Modules}} \\ \midrule
select         & -    & $\mathbf{d}^{out}$       & $concept, attr$      & find object named $concept$ &  \makecell{$\mathbf{e}_{i} = \mathcal{M}_{attr} (\mathbf{v}_{i})$, \\ $\mathbf{d}^{out}_{i} = \operatorname{sim} (\mathbf{e}_{i}, \mathbf{c}_{concept} )$}  \\ \hline
filter         & $\mathbf{d}^{in}$     & $\mathbf{d}^{out}$       & $concept, attr$          & from the input $\mathbf{d}^{in}$, find object whose attribute $attr$ is $concept$  & \makecell{$\mathbf{e}_{i} = \mathcal{M}_{attr} (\mathbf{v}_{i})$, \\ $\mathbf{d}^{res}_{i} = \operatorname{sim} (\mathbf{e}_{i}, \mathbf{c}_{concept} )$, \\ $\mathbf{d}^{out} = \operatorname{Merge}(\mathbf{d}^{in}, \mathbf{d}^{res})$} \\ \hline
relate\_o      & $\mathbf{d}^{in}_1, \mathbf{d}^{in}_2$       & $\mathbf{d}^{out}$       & $rel, rtype$ \tablefootnote{We assign each relationship into one of the three predefined relationship types ($rtype$), which are $\texttt{spatial}$ (\eg, to the left of, on top of, etc.), $\texttt{semantic}$ (\eg, wearing, holding, etc.) and $\texttt{spatial+semantic}$ (\eg, sitting on, looking at, etc.). } & find the object of $rel$ to $\mathbf{d}^{in}_2$ from $\mathbf{d}^{in}_1$  & \makecell{$\mathbf{e}_{ij} = \mathcal{M}^{r}_{rtype} (\mathbf{v}_{i}, \mathbf{v}_{j})$, \\ $\mathbf{mask}_{ij} = \operatorname{sim} (\mathbf{e}_{ij}, \mathbf{c}_{rel} )$, \\ $\mathbf{d}^{res}_{i} = \sum_{j=1}^{N} \mathbf{d}^{in}_{2j} \mathbf{mask}_{ij}$, \\ $\mathbf{d}^{out} = \operatorname{Merge}(\mathbf{d}^{in}_{1}, \mathbf{d}^{res})$}            \\ \hline
relate\_s      & $\mathbf{d}^{in}_1, \mathbf{d}^{in}_2$       & $\mathbf{d}^{out}$       &  $rel, rtype$         &  find the object of $rel$ to $\mathbf{d}^{in}_2$ from $\mathbf{d}^{in}_1$  & \makecell{$\mathbf{e}_{ij} = \mathcal{M}^{r}_{rtype} (\mathbf{v}_{j}, \mathbf{v}_{i})$, \\ $\mathbf{mask}_{ij} = \operatorname{sim} (\mathbf{e}_{ij}, \mathbf{c}_{rel} )$, \\ $\mathbf{d}^{res}_{i} = \sum_{j=1}^{N} \mathbf{d}^{in}_{2j} \mathbf{mask}_{ij}$, \\ $\mathbf{d}^{out} = \operatorname{Merge}(\mathbf{d}^{in}_{1}, \mathbf{d}^{res})$}  \\ \hline
relate\_ae     & $\mathbf{d}^{in}_1, \mathbf{d}^{in}_2$       & $\mathbf{d}^{out}$       & $attr$  & find the object from $\mathbf{d}^{in}_1$ that has the same $attr$ with $\mathbf{d}^{in}_{2}$ & \makecell{$\mathbf{e}_{i} = \mathcal{M}_{attr} (\mathbf{v}_{i})$, \\ $\mathbf{mask}_{ij} = \operatorname{sim}(\mathbf{e}_{i}, \mathbf{e}_{j})$, \\ $\mathbf{d}^{res}_{i} = \sum_{j=1}^{N} \mathbf{d}^{in}_{2j} \mathbf{mask}_{ij}$, \\ $\mathbf{d}^{out} = \operatorname{Merge}(\mathbf{d}^{in}_{1}, \mathbf{d}^{res})$} \\ \midrule
\multicolumn{6}{c}{{\bf Output Modules}} \\ \midrule
query          & $\mathbf{d}^{in}$       & $\mathbf{a}$       & $attr$   & query the attribute $attr$ of the given input $\mathbf{d}^{in}$ & \makecell{$\mathbf{e}_{i} = \mathcal{M}_{attr} (\mathbf{v}_{i})$, \\ $\mathbf{e} = \mathbf{d}^{in} \cdot [\mathbf{e}_{1}, \mathbf{e}_{2}, ..., \mathbf{e}_{N}]$, \\ $\mathbf{a}_{concept} = \operatorname{sim} (\mathbf{e}, \mathbf{c}_{concept})$, \\ $concept \in \mathcal{C}(attr)$ \tablefootnote{Here $\mathcal{C}(attr)$ represents the set of concepts of the given attribute $attr$.} } \\ \hline
query\_rel\_s  & $\mathbf{d}^{in}_1, \mathbf{d}^{in}_2$       & $\mathbf{a}$       &  $rtype$         & query the relationship between $\mathbf{d}^{in}_{1}$ (subject) and $\mathbf{d}^{in}_{2}$ (object) & \makecell{$\mathbf{e}_{ij} = \mathcal{M}^{r}_{rtype} (\mathbf{v}_{i}, \mathbf{v}_{j})$, \\ $\mathbf{e} = \sum_{i=1}^{N} \sum_{j=1}^{N} \mathbf{d}^{in}_{1i} \mathbf{e}_{ij} \mathbf{d}^{in}_{2j}$, \\ $\mathbf{a}_{rel} = \operatorname{sim} (\mathbf{e}, \mathbf{c}_{rel})$, \\ $rel \in \mathcal{C}(rtype)$ \tablefootnote{Here $\mathcal{C}(rtype)$ represents the set of relationships of the given relationship type $rtype$.} }\\ \hline
query\_rel\_o  & $\mathbf{d}^{in}_1, \mathbf{d}^{in}_2$       & $\mathbf{a}$       & $rtype$ & query the relationship between $\mathbf{d}^{in}_{1}$ (object) and $\mathbf{d}^{in}_{2}$ (subject) & $\mathbf{a} = \operatorname{query\_rel\_s}[rtype](\mathbf{d}^{in}_2, \mathbf{d}^{in}_1)$            \\ \hline
verify         & $\mathbf{d}^{in}$       & $a$       & $concept, attr$   & verify whether the attribute $attr$ of given input $\mathbf{d}^{in}$ is $concept$  & \makecell{$\mathbf{e}_{i} = \mathcal{M}_{attr} (\mathbf{v}_{i})$, \\ $\mathbf{e} = \mathbf{d}^{in} \cdot [\mathbf{e}_{1}, \mathbf{e}_{2}, ..., \mathbf{e}_{N}]$, \\ $a = \operatorname{sim} (\mathbf{e}, \mathbf{c}_{concept})$} \\ \hline
choose         & $\mathbf{d}^{in}_1, \mathbf{d}^{in}_2$       & $\mathbf{a}$       & $concept, attr$ & choose whether $\mathbf{d}^{in}_{1}$ or $\mathbf{d}^{in}_{2}$ is of $concept$ in specified attribute $attr$ & \makecell{$\mathbf{a}_{1} = \operatorname{verify}[attr, concept](\mathbf{d}^{in}_{1})$, \\ $\mathbf{a}_{2} = \operatorname{verify}[attr, concept](\mathbf{d}^{in}_{2})$} \\ \hline
verify\_rel\_s & $\mathbf{d}^{in}_1, \mathbf{d}^{in}_2$       & $a$       & $rel, rtype$ & verify whether $\mathbf{d}^{in}_1$ (subject) and $\mathbf{d}^{in}_2$ (object) are of relationship $rel$ & \makecell{$\mathbf{e}_{ij} = \mathcal{M}^{r}_{rtype} (\mathbf{v}_{i}, \mathbf{v}_{j})$, \\ $\mathbf{e} = \sum_{i=1}^{N} \sum_{j=1}^{N} \mathbf{d}^{in}_{1i} \mathbf{e}_{ij} \mathbf{d}^{in}_{2j}$, \\ $a = \operatorname{sim} (\mathbf{e}, \mathbf{c}_{rel})$ } \\ \hline
verify\_rel\_o & $\mathbf{d}^{in}_1, \mathbf{d}^{in}_2$       & $a$       & $rel, rtype$   &  verify whether $\mathbf{d}^{in}_1$ (object) and $\mathbf{d}^{in}_2$ (subject) are of relationship $rel$ & $a = \operatorname{verify\_rel\_s}[rtype](\mathbf{d}^{in}_2, \mathbf{d}^{in}_1)$            \\ \hline
same           & $\mathbf{d}^{in}$       & $a$       &  $attr$  & whether objects in $\mathbf{d}^{in}$ have the same $attr$ & \makecell{$\mathbf{e}_{i} = \mathcal{M}_{attr} (\mathbf{v}_{i})$, \\ $\mathbf{e}=\frac{1}{N} \sum_{i=1}^{N} \mathbf{e}_{i}$, \\ $a = \sum_{i=1}^{N} \mathbf{d}^{in}_{i} \operatorname{sim}(\mathbf{e}_{i}, \mathbf{e})$} \\ \hline
query\_ae      & $\mathbf{d}^{in}_1, \mathbf{d}^{in}_2$       & $a$       &  $attr$ & whether $\mathbf{d}^{in}_{1}$ and $\mathbf{d}^{in}_{2}$ have the same $attr$ & \makecell{$\mathbf{e}_{i} = \mathcal{M}_{attr} (\mathbf{v}_{i})$, \\ $\mathbf{e}^{1}=\sum_{i=1}^{N} \mathbf{d}^{in}_{1i} \mathbf{e}_{i}$, \\ $\mathbf{e}^{2}=\sum_{i=1}^{N} \mathbf{d}^{in}_{2i} \mathbf{e}_{i}$, \\ $a = \operatorname{sim}(\mathbf{e}^{1}, \mathbf{e}^{2})$} \\ \hline
common         & $\mathbf{d}^{in}_1, \mathbf{d}^{in}_2$       & $\mathbf{a}$       &  -  & what attribute do $\mathbf{d}^{in}_1$ and $\mathbf{d}^{in}_2$ share & \makecell{For all possible $attr$s: \\$\mathbf{a}_{attr} = \operatorname{query\_ae}[attr](\mathbf{d}^{in}_{1}, \mathbf{d}^{in}_{2})$}            \\ \hline
exist          & $\mathbf{d}^{in}$       & $a$       & -          & whether object $\mathbf{d}^{in}$ exists   & $a = \operatorname{max}(\mathbf{d}^{in})$  \\ \hline
intersect      & $a^{in}_{1}, a^{in}_{2}$       & $a$       &  -         &  whether both $a^{in}_{1}$ AND $a^{in}_{2}$ are true   & ${a} = \operatorname{min}({a}^{in}_{1}, {a}^{in}_{2})$            \\ \hline
union          & ${a}^{in}_{1}, {a}^{in}_{2}$       & ${a}$       &  -         &  whether ${a}^{in}_{1}$ OR ${a}^{in}_{2}$ is true                            &  ${a} = \operatorname{max}({a}^{in}_{1}, {a}^{in}_{2})$           \\ \bottomrule 
\end{tabular}}
\end{table}

\end{document}